\documentclass[letterpaper]{article} 
\usepackage{aaai24}  
\usepackage{times}  
\usepackage{helvet}  
\usepackage{courier}  
\usepackage[hyphens]{url}  
\usepackage{graphicx} 
\usepackage{natbib}  
\usepackage{tabularx}
\usepackage{caption} 
\usepackage{bm}
\usepackage{booktabs}
\usepackage{multirow}
\usepackage{array}
\usepackage{amsmath}
\usepackage{amsfonts}
\usepackage[nameinlink]{cleveref}

\title{Faithful Knowledge Distillation}
\author {
    Tom A. Lamb,\textsuperscript{\rm 1} 
    Rudy Brunel,\textsuperscript{\rm 2} 
    Krishnamurthy (Dj) Dvijotham,\textsuperscript{\rm 2} 
    M. Pawan Kumar,\textsuperscript{\rm 2} 
    Philip H.S. Torr,\textsuperscript{\rm 3} 
    Francisco Eiras\textsuperscript{\rm 3}
}
\affiliations{
    \textsuperscript{\rm 1}University of Edinburgh,
    \textsuperscript{\rm 2}Google DeepMind, 
    \textsuperscript{\rm 3}University of Oxford,
}

\newtheorem{definition}{Definition}
\newcommand{\lb}{l_b}
\newcommand{\ub}{u_b}
\DeclareMathOperator*{\argmax}{arg\,max}

\newcommand{\indicator}[1]{\mathbf{1}_{\{#1\}}}

\begin{document}
\maketitle

\begin{abstract}
Knowledge distillation (KD) has received much attention due to its success in compressing networks to allow for their deployment in resource-constrained systems. While the problem of adversarial robustness has been studied before in the KD setting, previous works overlook what we term the \textit{relative calibration} of the student network with respect to its teacher in terms of soft confidences. In particular, we focus on two crucial questions with regard to a teacher-student pair: (i) do the teacher and student disagree at points close to correctly classified dataset examples, and (ii) is the distilled student as confident as the teacher around dataset examples?
These are critical questions when considering the deployment of a smaller student network trained from a robust teacher within a safety-critical setting. To address these questions, we introduce a \textit{faithful imitation} framework to discuss the relative calibration of confidences and provide empirical and certified methods to evaluate the relative calibration of a student w.r.t. its teacher. Further, to verifiably align the relative calibration incentives of the student to those of its teacher, we introduce \textit{faithful distillation}.
Our experiments on the MNIST, Fashion-MNIST and CIFAR-10 datasets demonstrate the need for such an analysis and the advantages of the increased verifiability of faithful distillation over alternative adversarial distillation methods.
\end{abstract}

\section{Introduction}\label{sec:intro}
The state-of-the-art performance of deep neural networks in various application areas has recently been fuelled by significant increases in their capacity~\citep{brown2020language,bommasani2021opportunities}. However, the increase in the size of networks has led to deployment issues for resource-constrained systems~\citep{gou2021knowledge,mishra2017apprentice} such as self-driving cars and small medical devices~\citep{wang2018private}. Simply deploying smaller versions of these networks, trained as usual, tends to hurt performance.

\begin{figure}[t]
\includegraphics[width=0.95\linewidth]{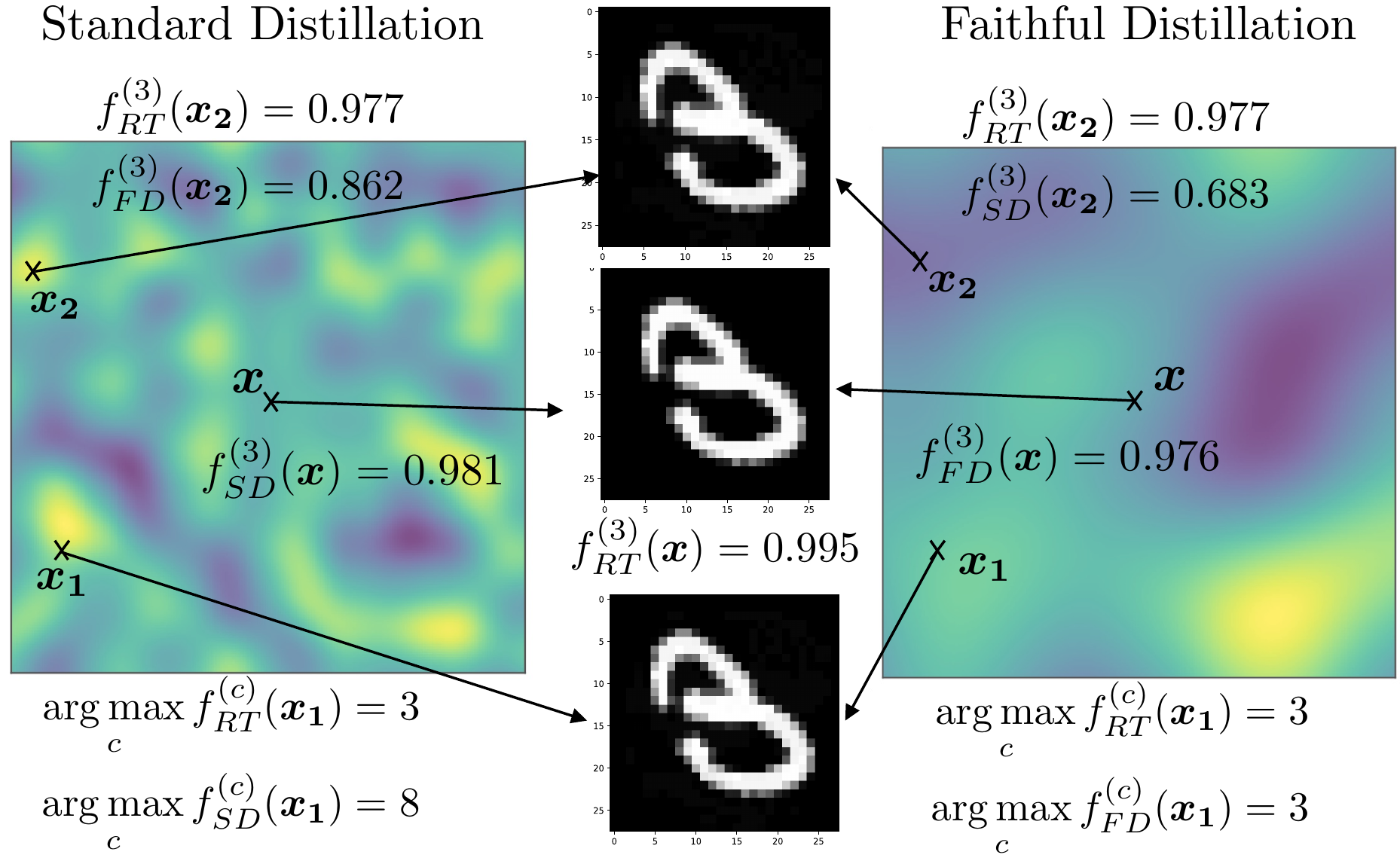}
\caption{Faithful distillation - a heat map showing the maximum difference (specifically the $\ell_{\infty}$-norm) of the soft confidence outputs of a robust teacher model ($f_{RT}$) and its two distilled students, a student trained using standard KD ($f_{SD}$), and a student trained using faithful distillation ($f_{FD})$ within an $\ell_{\infty}$-ball surrounding an example $\bm{x}$ from the MNIST dataset. In particular, we observe two problematic scenarios for $f_{SD}$: an adversarial disagreement example at $\bm{x}_2$, and poor relative calibration in terms of confidences, that is, large differences in confidence outputs of the two networks, at $\bm{x}_1$. We can see for $f_{FD}$, the heat map transitions are a lot smoother, indicating smaller differences in confidences with its teacher. This implies that $f_{FD}$ more closely imitates its teacher than $f_{SD}$. Overall, we observe that faithful distillation improves upon the two problems observed with $f_{SD}$.}
\label{fig:motivation}
\end{figure}

Knowledge distillation (KD) helps to deal with this problem by distilling knowledge from a large, high-performing, expensive-to-run neural network into a smaller network~\citep{gou2021knowledge, hinton2015distilling,tung2019similarity}. This has been shown to improve the performance of smaller networks compared to standard training~\citep{hinton2015distilling}, bringing the performance of larger networks to smaller networks which tend to be more efficient to run. However, students trained using standard KD are vulnerable to \textit{adversarial attacks}, that is, to small input perturbations which cause networks to change their correct classification~\citep{robuststudentgou, goldblum2020adversarially}.

To mitigate this issue, previous works have studied the robustness of teacher and student networks separately, in particular, that distilling knowledge from a robust teacher improves the robustness of the distilled student~\citep{goldblum2020adversarially}. However, an important question arises: if we have a robust teacher, \textit{can we find dataset examples where the teacher and student agree, yet under small perturbations, then disagree}? Moreover, consider the case where the confidence of the deployed network will be used to make decisions - \textit{is the distilled student as confident as the teacher in and around dataset examples}? As \Cref{fig:motivation} shows, small perturbations of an image can cause significant disagreements in classification, which we refer to as \textit{adversarial disagreement examples}. Moreover, such small perturbations can lead to major differences in confidences between a teacher and its student, what we refer to as poor \textit{relative calibration} of confidences. This motivates the need for (i) a framework to evaluate the robustness and confidence of the student with respect to the teacher and (ii) a training method that allows us to obtain better students with respect to that framework.

Our contributions are fourfold:
\begin{itemize}
    \item we define the concept of a \textit{faithful imitator} to discuss and bound the difference in confidences of a teacher and its student;
    \item we introduce novel empirical and verified methods to investigate and compute these bounds;
    \item we provide a \textit{faithful distillation loss} that we show is a viable alternative method of adversarial distillation, which in particular allows us to train students which are verifiably more aligned with their teacher in terms of confidences, and
    \item we demonstrate the capabilities of our framework on the MNIST~\citep{lecun1998gradient}, Fashion-MNIST~\citep{xiao2017fashion} and CIFAR-10~\citep{krizhevsky2010convolutional} datasets.
\end{itemize}

\section{Background and Related Work}
 Throughout this paper, we assume that we are dealing with a $C$-class classification problem and that $(\bm{x},y) \sim \mathcal{D}$ is an example from a data distribution $\mathcal{D}$, with $\bm{x} \in \mathbb{R}^n$ an input vector and $y\in \{1,\dots,C\}$ its associated class. 

\subsection{Knowledge Distillation}
\label{sec: KD}
Knowledge distillation (KD) is the process of transferring information and representations from a larger teacher network, $f_t: \mathbb{R}^{n} \to [0, 1]^{C}$, to a usually significantly smaller student network, $f_s: \mathbb{R}^{n} \to [0, 1]^{C}$, with the standard aim of improving the performance of the student over regular training. A popular form of KD which we refer to as \textit{standard distillation }was introduced by \citet[SD]{hinton2015distilling}. SD encourages the clean outputs of a student network to match the clean outputs of its teacher whilst also balancing the accuracy of the student's predictions.  

\subsection{Adversarial Attacks and Training}
\label{sec: adversarial attacks and training}
Adversarial examples are dataset examples that a network correctly classifies that can be deliberately perturbed to cause the network to change its classification \citep{goodfellow2014explaining}. Several white-box \citep{goodfellow2014explaining, dong2018boosting, madry2017towards} and black box-techniques \citep{narodytska2017simple, ilyas2018black, li2019nattack} have been developed to generate adversarial examples. One such method utilises PGD attacks which make use of projective gradient ascent to maximise the training loss $\mathcal{L}$ of a network within a constrained domain $\mathcal{S}$ (usually an $\ell_p$-ball surrounding a given example that we wish to attack, typically for $p=\infty$) \citep{madry2017towards}.

Several methods to defend against adversarial attacks have been introduced~\citep{song2017pixeldefend, samangouei2018defense, guo2017countering}. Such methods include adversarial training~\citep{goodfellow2014explaining, madry2017towards, na2017cascade}, which aims to train individual networks to be more robust to adversarial attacks. \citet{madry2017towards} framed the problem of adversarial training as a saddle point problem where they make use of PGD as described above to tackle the inner maximisation step of the problem.

\subsection{Adversarial Distillation Methods}
\label{sec: adversarial distillation}
Related to adversarial training of individual networks, alternative methods of knowledge distillation building upon SD have been proposed to encourage distilled student networks to be more robust to adversarial attacks.

One such method is Adversarially Robust Distillation \citep[ARD]{goldblum2020adversarially}, which is a modification of the SD method discussed in \Cref{sec: KD}. ARD is implemented so that the distilled student agrees with its teacher's unperturbed confidences even when the student itself is perturbed~\citep{goldblum2020adversarially}. This allows ARD training to be used to not only distil knowledge but also robustness from a teacher network. The authors found that using ARD can experimentally produce more robust student networks that can even beat the performance in terms of robust accuracy of networks trained adversarially with the same network architecture~\citep{goldblum2020adversarially}.

Robust Soft Label Adversarial Distillation (RSLAD) is an adaption of ARD that replaces the hard labels within the loss with soft labels given by the softmax outputs of the teacher~\citep{zi2021revisiting}. \citet{zi2021revisiting} found that RSLAD empirically produced more robust students than ARD with respect to both PGD attacks and even more sophisticated attacks such as AutoAttack \citep{croce2020reliable}.

Introspective adversarial distillation \citep[IAD]{zhu2021reliable} is a method of distillation where students only partially trust their teacher during training, only when the student is confident that the teacher is reliable on a particular perturbed training example. IAD is implemented as a version of ARD that incorporates the uncertainty of a teacher during the distilling of a student. Despite showing success in producing robust student networks, \citet{zi2021revisiting}, observe that the outputs of distilled IAD students are less aligned to their teacher's outputs than students distilled using RSLAD.

\section{Motivation}
\label{sec:motivation}
We motivate the need for a method of \textit{faithful distillation} through the examples illustrated in \Cref{fig:motivation}. Given a robust teacher network, we distil a student using the SD loss introduced by \citet{hinton2015distilling}.

Let us consider an example of an image $\bm{x}$ from the MNIST dataset~\citep{lecun1998gradient}. Both networks correctly classify the image $\bm{x}$ as a $3$ with similar, high confidence levels. Let us now investigate perturbations of $\bm{x}$ in a neighbourhood of $\bm{x}$, specifically in an $\ell_{\infty}$-ball of radius $\epsilon = 0.1$.

We first find $\bm{x}_1$, a perturbation of the image $\bm{x}$, where the teacher classifies the perturbed image as a 3. However, the student classifies the perturbed image as an 8, despite the minimal difference between $\bm{x}_1$ and the original image $\bm{x}$. We will refer to an example where the student and teacher initially agree on a prediction but, under perturbation, the networks disagree with one another as an \textit{adversarial disagreement example}. Formally, we define it as below:

\begin{definition}[Adversarial Disagreement Example] \label{def:distillation_adversarial_example}
For an input $\bm{x}\in \mathbb{R}^n$ and teacher and student networks, $f_t: \mathbb{R}^n \to [0,1]^C$ and $f_s: \mathbb{R}^n \to [0,1]^C$ respectively, we say that $\bm{x}$ is an adversarial disagreement example for a given $\epsilon > 0$ if there exists an $\bm{x}' \in B_{\epsilon}(\bm{x})$ such that 
\begin{equation}
c_t(\bm{x}) = c_s(\bm{x}) \text{ and } c_t(\bm{x}') \neq c_s(\bm{x}'),
\end{equation}
for $c_t(\bm{x}) = \argmax_c f^c_t(\bm{x})$, $c_s(\bm{x}) = \argmax_cf^c_s(\bm{x})$ and where $B_{\epsilon}(\bm{x})$ denotes an $\ell_p$-ball of radius $\epsilon$ centred at $\bm{x}$.
\end{definition}

We can view such examples as perturbations that cause an originally in agreement teacher-student pair to disagree. The quantity of such examples within a test set gives a measure of the robustness of the distillation process itself. 

Next, we find $\bm{x}_2$, another perturbation of the image $\bm{x}$, where the teacher and student classify $\bm{x}_2$ as a $3$, but with very different confidences in their predictions. Here, we would say that \textit{the student is not relatively well calibrated in terms of confidence} to its teacher. This poses significant problems in the confident deployment of student networks in safety-critical environments where confidences are used in decision-making. In such a scenario, not only are correct classifications important, but also the confidence in said classification. Here, we see that even a small perturbation can result in wildly different confidences, which implies that such a deployed student poorly imitates and is not relatively well-calibrated to its teacher.

The above motivates the need for a framework to empirically investigate and verify how relatively well-calibrated a student is to its teacher in terms of their confidences. Moreover, it calls for a form of \textit{faithful distillation} that addresses the issues discussed above as measured within this new framework. Therefore, we present the faithful imitation framework in 
\Cref{sec:faithful_imitator} and a loss with the aim of producing verifiably more relatively calibrated students in \Cref{sec:faithful_distillation}.

\section{Faithful Imitator For Knowledge Distillation}
\label{sec:faithful_imitator}
In this section, we provide an evaluation framework based on the concept of faithful imitation, along with methods to compute lower and upper bounds based on it.

Assume we have two processes defined by the functions $f: \mathbb{R}^n \to \mathbb{R}^d$ and $\hat{f}: \mathbb{R}^n \to \mathbb{R}^d$, where the goal of $\hat{f}$ is to imitate the output of $f$. We define a \textit{faithful imitator} as follows.

\begin{definition}[Faithful Imitator] \label{def:faithful-imitator}
We say that $\hat{f}: \mathbb{R}^n \to \mathbb{R}^d$ is an $(\epsilon, \delta)$- faithful imitation of $f: \mathbb{R}^n \to \mathbb{R}^d$ around $\bm{x}_0 \in \mathbb{R}^n$ if
\begin{equation}
d_{f}(f(\bm{x}), \hat{f}(\bm{x})) \leq \delta, \quad \forall \bm{x} \in B_{\epsilon}(\bm{x}_0),
\end{equation}
where $B_{\epsilon}(\bm{x}_0) = \{\bm{x}' \in \mathbb{R}^n \; |\; d_{\bm{x}}(\bm{x}_0, \bm{x}') \leq \epsilon \}$, $d_{f}: \mathbb{R}^d \times \mathbb{R}^d \to \mathbb{R}_{\geq 0}$ is a chosen metric function in the output space, and $d_{\bm{x}}: \mathbb{R}^n \times \mathbb{R}^n \to \mathbb{R}_{\geq 0}$ is a metric function in the input space. We refer to any $\delta$ that bounds $d_{f}(f(\bm{x}'), \hat{f}(\bm{x}'))$ as a faithfulness bound for a given $\epsilon$.
\end{definition}

Intuitively, this definition holds if for an $\epsilon$-neighbourhood defined by a function $d_{\bm{x}}$ around an input $\bm{x}_0$, the outputs of the imitator $\hat{f}$ and the original process $f$ are \textit{similar} - up to a difference of $\delta$ - with respect to an output metric $d_{f}$.

Within the context of knowledge distillation, we have a student network, $f_s$, that is trying to imitate the output of a teacher network, $f_t$. We can use the definition of a faithful imitator as a principled way of reasoning about the \textit{relative calibration} of a student network with respect to its teacher in terms of confidences. \textit{By relative calibration, we refer to the similarity in the confidence outputs of the two networks.} For this purpose, for a given $\epsilon$, we are interested in computing the tightest $\delta$ that satisfies \Cref{def:faithful-imitator}. In the setting of multi-class classification, we follow the robustness literature and define $d_{\bm{x}}$ and $d_{f}$ to be the metrics induced from the $\ell_\infty$-norm~\citep{madry2017towards}.

For the sake of simplicity of our framework, within a $C$-class knowledge distillation setting, we assume $f_t: \mathbb{R}^{d_0} \to [0, 1]^{C}$ and $f_s: \mathbb{R}^{d_0} \to [0, 1]^{C}$ to be $L_t$ and $L_s$ fully connected neural networks, respectively, whose outputs are then normalised by a softmax function, $\sigma$, to yield the class confidences. That is, $f_t(\bm{x}) = \sigma(g_t(\bm{x}))$ and $f_s(\bm{x}) = \sigma(g_s(\bm{x}))$, where $g_t$ and $g_s$ are fully connected networks.

Formally, we define an $L$-layer fully connected neural network as a function $g: \mathbb{R}^{d_0} \to \mathbb{R}^{d_L}$, such that for an input $\bm{x}\in \mathbb{R}^{d_0}$, $g(\bm{x}) = \bm{z}^{(L)}(\bm{x}) = \bm{W}^{(L)} \bm{z}^{(L-1)}(\bm{x}) + \bm{b}^{(L)}$, where $\bm{z}^{(k)}(\bm{x}) = \bm{W}^{(k)} \phi(\bm{z}^{(k-1)}(\bm{x})) + \bm{b}^{(k)}$ for $k \in \{1,\dots,L-1\}$, and $\bm{z}^{(0)}(\bm{x}) = \bm{x}$, in which $\bm{W}^{(k)} \in \mathbb{R}^{d_k \times d_{k-1}}$ and $\bm{b}^{(k)} \in \mathbb{R}^{d_k}$ are the weight and bias associated with the $k$-th layer of the network, and $\phi: \mathbb{R} \to \mathbb{R}$ is the element-wise activation function.

The computation of the optimal $\delta$ that satisfied \Cref{def:faithful-imitator} can be written in the form shown in \Cref{eq:optimization_setup} below:
\begin{equation}
\small
\label{eq:optimization_setup}
\begin{split}
\max_{\bm{x} \in  B_{\epsilon}(\bm{x}_0)}\; & \lVert f_t(\bm{x}) - f_s(\bm{x}) \rVert_{\infty}
= \lVert \sigma(g_t(\bm{x})) - \sigma(g_s(\bm{x})) \rVert_{\infty} \\
\text{s.t.}\quad & g_t(\bm{x}) = \bm{W}_t^{(L_t)} \bm{z}_t^{(L_t-1)}(\bm{x}) + \bm{b}_t^{(L_t)}\\
& \bm{z}_t^{(k_t)}(\bm{x}) = \bm{W}_t^{(k_t)} \phi \left(\bm{z}_t^{(k_t-1)}(\bm{x}) \right) + \bm{b}_t^{(k_t)} \\
& \bm{z}_t^{(0)}(\bm{x}) = \bm{x} \\
& \bm{g}_s(\bm{x}) = \bm{W}_s^{(L_s)} \bm{z}_s^{(L_s-1)}(\bm{x}) + \bm{b}_s^{(L_s)}\\
& \bm{z}_s^{(k_s)}(\bm{x}) = \bm{W}_s^{(k_s)} \phi \left(\bm{z}_s^{(k_s-1)}(\bm{x}) \right) + \bm{b}_s^{(k_s)} \\
& \bm{z}_s^{(0)}(\bm{x}) = \bm{x} \\
& k_t \in \{1,\dots,L_t-1\} \\ 
& k_s \in \{1,\dots,L_s-1\}.
\end{split}
\end{equation}

When solved to optimality for $\bm{x}\in \mathbb{B}_\epsilon(\bm{x}_0)$, the $\delta^*$ that maximises the differences between the confidence outputs of the two networks implies $f_s$ is a $(\epsilon, \delta^*)$-faithful imitator of $f_t$ around $\bm{x}_0$, meaning we can \textit{deploy $f_s$ with a guarantee that the confidence outputs of the student are similar to the teacher's outputs}.

The issue with solving \Cref{eq:optimization_setup} to optimality is that this is a general non-linear optimisation problem. In the following sections, we approach the computation of $\delta^*$ in two different ways: using an empirical, best-effort optimiser that gives us a lower bound on $\delta^*$ and by over-approximating and linearising the problem in a similar fashion to~\citet{zhang2018efficient} obtaining an upper bound on $\delta^*$. The use of upper and lower bounds on the solution to \Cref{eq:optimization_setup} provides us with a tractable way of evaluating the faithfulness, that is, the degree of imitation in terms of confidences, of a student network to its teacher.

\subsection{Empirical Lower Bounds on $\delta^{*}$} \label{Empirical Lower Bounds}
We can empirically obtain a lower bound on the solution to \Cref{eq:optimization_setup} by using PGD attacks discussed in \Cref{sec: adversarial attacks and training} to maximise $\lVert  f_t(\bm{x}) - f_s(\bm{x}) \rVert_{\infty}.$ By using PGD to maximise the difference in confidences between the student and teacher networks within the $\ell_\infty$-ball around a given dataset point $\bm{x}_0$, we obtain a best-effort lower bound on $\delta^*$. This is useful for empirically investigating how large the difference in confidences of the teacher and student can be, but \textit{gives us no guarantee on the maximum possible difference within an $\epsilon$-ball surrounding the given image $\bm{x}$}, and therefore cannot be considered a faithfulness bound. 

\subsection{Faithfulness Upper Bounds on $\delta^*$} \label{Faithfulness Upper Bounds}
To compute faithfulness bounds as per \Cref{def:faithful-imitator}, we over-approximate the solution of \Cref{eq:optimization_setup} by relaxing the problem using a linear formulation. We achieve this by (i) relaxing the non-linear activation functions $\phi$ at each layer using linear lower and upper bounds and (ii) relaxing the softmax $\sigma$ that yields $f_t(\bm{x}) = \sigma(g_t(\bm{x}))$ (and similarly for the student network).

Assuming that $\bm{z}^{(k)}_L \leq \bm{z}^{(k)} \leq \bm{z}^{(k)}_U$ for a given $k$ and an activation function $\phi$, we compute the parameters $\bm{\alpha}_{L}^{(k)}(\bm{z}^{(k)} + \bm{\beta}^{(k)}_L) \leq \phi(\bm{z}^{(k)}) \leq \bm{\alpha}^{(k)}_U (\bm{z}^{(k)} + \bm{\beta}^{(k)}_U)$. For $\phi$ a ReLU activation, we can obtain the parameters for the relaxations provided in \citet{ehlers2017formal} and \citet{zhang2018efficient}.

For an input vector $\bm{z} \in \mathbb{R}^C$, the softmax operator outputs a vector where the $i$-th component is defined as $\sigma(\bm{z})_i = \exp(z_i) / \sum_j \exp(z_j)$. Note that the softmax can be thus be written as $\sigma(\bm{z})_i = 1/( \sum_{j \neq i} \exp(z_j - z_i) + 1)$. Using this, we bound the $i$-th component of the softmax activation by first bounding the difference between logits. Specifically, given upper and lower bounds on the logits $ z_i^{\lb} \leq z_i \leq z_i^{\ub} $ we compute the following over this domain: 
\begin{equation}
\small
d^{lb}_{i,j} = \min_{\bm{z}} (z_j - z_i) \; \; , \; \; d^{\ub}_{i, j} = \max_{\bm{z} } (z_j - z_i) ,
\end{equation}
giving $d^{\lb}_{i,j} \leq z_j - z_i \leq d^{\ub}_{i,j}$ over inputs $\bm{z}$ whose components satisfy $ z_i^{\lb} \leq z_i \leq z_i^{\ub} $. 

To obtain bounds on the $i$-th component of the softmax for each network, we propagate the logit difference bounds through the softmax function:
\begin{equation}
\small
\sigma_i^{\lb}  = \frac{1}{1+ \sum_{j \neq i}\exp{\left(d_{i,j}^{ \ub}\right)}} \; , \; 
\sigma_i^{\ub}  = \frac{1}{1+ \sum_{j \neq i}\exp{\left(d_{i,j}^{ \lb}\right)}}.
\end{equation}
This gives us $\sigma_i^{\lb} \leq \sigma (\bm{z})_i \leq \sigma_i^{\ub}$ for $ z_i^{\lb} \leq z_i \leq z_i^{\ub} $.

We apply the bounding of the softmax activation to $f_t$ and $f_s$ by using bounds on their logits, $z_{t, i}^{(L_t)}$ and $z_{s, i}^{(L_s)}$ respectively, to obtain upper and lower bounds of their soft outputs, $\bm{\sigma}^{\ub}(f_t(\bm{x}))$ and $\bm{\sigma}^{\lb}(f_s(\bm{x}))$. We then compute a faithfulness upper on bound $\delta^*$ given by:
\begin{align}
\small
\begin{split}
\max_{\bm{x}_\in B_\epsilon(\bm{x}_0)} &  \lVert f_t(\bm{x}) - f_s(\bm{x}) \rVert_{\infty} \\
\leq \max \Bigl( & \left\Vert \bm{\sigma}^{\ub}(f_t(\bm{x})) - \bm{\sigma}^{\lb}(f_s(\bm{x})) \right\rVert_{\infty}  ,  \\
& \left\lVert \bm{\sigma}^{\ub}(f_s(\bm{x})) - \bm{\sigma}^{\lb}(f_t(\bm{x}))  \right\rVert_{\infty} \Bigl).
\end{split}
\end{align}
For small enough $f_t$ and $f_s$, using a MILP solver directly, such as Gurobi~\citep{gurobi}, yields tighter bounds following the description above within reasonable runtimes compared to alternative bound propagation methods such as CROWN \citep{zhang2018efficient}.

\paragraph{Extension to Convolutional Neural Networks.} One can extend the formulation introduced in \Cref{eq:optimization_setup} and the bounding method introduced in \Cref{Faithfulness Upper Bounds} to networks such as convolutional neural networks. In particular, since a convolution is a linear operation, this can be bounded similarly to a standard fully-connected layer. We can further bound max-pooling layers as follows: given upper and lower bounds on the input neurons of a max pooling layer $\bm{z}^{(k)}_L \leq \bm{z}^{(k)} \leq \bm{z}^{(k)}_U$, we obtain upper and lower bounds on the max pooling layer by simply propagating these bounds through the max-pooling function $\text{MaxPool}$ so that $\text{MaxPool}(\bm{z}^{(k)}_L) \leq \text{MaxPool}(\bm{z}^{(k)}) \leq \text{MaxPool}(\bm{z}^{(k)}_U)$.

\paragraph{Connection to Model Calibration}
Our proposed method of producing faithfulness bounds provides an upper bound on the relative calibration of teacher and student networks, that is, the difference in confidences. Therefore, a corollary of the introduced framework above is that if a teacher is robust and is generally well-calibrated on a dataset \citep{bella2010calibration}, a distilled student that is a faithful imitator (i.e. achieves low empirical and faithfulness bounds) of its teacher will also be well-calibrated.

\section{Faithful Distillation}
\label{sec:faithful_distillation}
Following our discussion in \Cref{sec:motivation}, we introduce a new form of distillation that we will refer to as \textit{Faithful Distillation} (FD), which is defined by the following loss function:
\begin{align} \label{FD loss}
\small
\begin{split}
\mathcal{L}_{FD}(\Theta)  = \alpha T^2  & \mathcal{L}_{KL}  \left(f_{s}(\bm{x} +\bm{\delta}_0  ;  T) , f_{t}(\bm{x} + \bm{\delta}_0  ;  T) \right) \\
&+ (1-\alpha)  \mathcal{L}_{CE} \left(f_{s}(\bm{x}  ;  1)  ,  y \right), 
\end{split}
\end{align}
where $\bm{\delta}_0 = \argmax_{\lVert \bm{\delta} \rVert_{\infty} \leq \epsilon}  \mathcal{L}_{KL} \left(f_{s}(\bm{x} + \bm{\delta})  ,  f_{t}(\bm{x} + \bm{\delta}) \right)$.

This loss is similar in structure to the ARD, RSLAD and IAD losses introduced in \Cref{sec: adversarial distillation}. However, all three of these methods produce adversaries during training by only perturbing the student network. On the other hand, FD generates adversaries during training by perturbing both the teacher and student networks. In doing this, we aim to better match the teacher and its student's confidences even when both models are perturbed. This should encourage the student network to match its teacher's confidences even under perturbation, leading to a more relatively calibrated or better-imitating student.

It is worth noting that both FD and RSLAD should achieve comparable performance since they should produce very similar bounds in the limit. This is because $f_t(x + \delta_0) \to f_t(x)$ as the teacher becomes increasingly robust during adversarial training. However, in \Cref{sec: experiments}, we observe that on MNIST and Fashion-MNIST, FD seems to produce more \textit{verifiably relatively calibrated teacher-student pairs}, which importantly provides a sound certificate for downstream tasks. Moreover, the \textit{increased verifiability is necessary in order to complement the less reliable empirical measures} of faithfulness such as $\textsc{EmpLB}$ introduced above. We discuss this topic in more depth in \Cref{sec: FD vs RSLAD}.

\section{Experiments}
\label{sec: experiments}
To demonstrate the framework of faithful imitation and to evaluate FD, we conduct experiments on the MNIST, F-MNIST and CIFAR-10 datasets. For comparison of FD, we distil student models on each dataset using different methods of distillation. Following the observation of \citet{zi2021revisiting} discussed in \Cref{sec: adversarial distillation}, and due to ARD's similarity in loss form to FD, we investigate SD, ARD and RSLAD in particular. 

Following the description in \Cref{sec:faithful_imitator}, on MNIST and F-MNIST, we train teacher and student networks which are fully connected, feed-forward networks utilising ReLU activations, rendering the student networks approximately 37.7\% and 52.5\% smaller than their respective teacher networks. Specific details of model architectures and training can be found in \Cref{MNIST training} and \Cref{Fashion - MNIST training}, respectively. 

For CIFAR-10, we train convolutional neural networks comprising of convolutional, max-pooling, ReLU activations, and fully-connected layers, resulting in student networks that are approximately 48\% smaller than their teachers. Specific network architectures for CIFAR-10, in particular, are detailed in \Cref{tab:CIFAR-10 network_architectures}, with training details shown in \cref{CIFAR training}.

Adversarial training and distillation were both conducted using 10 iterations of PGD augmentations \citep{madry2017towards}, with $\epsilon=0.2$ and a step size of $0.05$, for MNIST, and $\epsilon=8/255$ and a step size of $2/255$ for F-MNIST and CIFAR-10. 

We begin by training robust teacher networks on each dataset, which we denote by $f_{RT}$ as we would like to investigate and produce robust and relatively well-calibrated students distilled from these robust teachers \citep{goldblum2020adversarially}. These teachers attain test set accuracies of 97.88\%, 88.23\%, and 63.09\% for MNIST, F-MNIST, and CIFAR-10, respectively. 

From these robust teachers, we train four distinct student networks via distillation on each dataset using SD, ARD, RSLAD and FD. Specific training details and training outcomes can be found in \Cref{MNIST training}, \Cref{Fashion - MNIST training}, and \Cref{CIFAR training}, respectively. Additionally, we perform an ablation study on CIFAR-10 to investigate the effect of changing the loss used for FD training following \Cref{FD loss}. The results of this study can be found in \Cref{Ablation}.

\paragraph{Measuring Robustness and Faithfulness.}
To compare the performance of the four different students with respect to their robust teacher $f_{RT}$ on each dataset, we are interested in measuring the robustness and faithfulness of the models

For robustness, we analyse both the teacher and student networks individually by computing each model's robust accuracy using 50-step PGD attacks at different $\epsilon$ values with a step size of $2.5 \epsilon / 50$ \citep{madry2017towards,goldblum2020adversarially}. 

Further, to understand the additional errors the students make compared to their robust teacher - as per the motivation in \Cref{sec:motivation} and \Cref{def:distillation_adversarial_example}- we introduce the concept of distillation agreement, which is simply the percentage of examples from a dataset that do not cause disagreements between teacher and student under perturbation.

\begin{definition}[Distillation Agreement] \label{Distillation Agreement}
For a dataset $\mathcal{D}$, teacher and student networks, $f_t$ and $f_s$ respectively, and $\epsilon > 0$, we define the set of original agreements, $\mathcal{S}_{OA} \subset \mathcal{D}$, and the set of adversarial disagreement examples, $\mathcal{S}_{ADE, \epsilon} \subset \mathcal{D}$, as
\begin{align}
\small
\begin{split}
&\mathcal{S}_{OA}^{t, s} = \{ \bm{x} \in \mathcal{D}:\, c_s(\bm{x}) = c_t(\bm{x}) \},\\
&\mathcal{S}_{ADE, \epsilon}^{t, s} = \{ \bm{x} \in \mathcal{S}_{OA}^{t,s} : \exists \bm{x}' \in B_{\epsilon}(\bm{x})  \text{ s.t. } c_t(\bm{x}') = c_s(\bm{x}') \},
\end{split}
\end{align}
for $c_t(\bm{x}) = \argmax_c\,f^c_t(\bm{x})$, $c_s(\bm{x}) = \argmax_cf^c_s(\bm{x})$.
We then define the \textit{distillation agreement}, $\mathcal{A}^t_{ \epsilon}(f_s)$, of the student $f_s$ w.r.t. its teacher $f_t$ on $\mathcal{D}$ for a given $\epsilon >0$ as:
\begin{equation}
\small
\mathcal{A}^t_{ \epsilon}(f_s) = \frac{1}{{\lvert \mathcal{D}\rvert}}\sum_{\bm{x} \in \mathcal{D}} \left( \indicator{\mathcal{S}^{t,s}_{OA}}(\bm{x}) - \indicator{\mathcal{S}^{t,s}_{ADE, \epsilon}}(\bm{x}) \right), 
\end{equation}
where as usual, $\lvert \mathcal{D}\rvert$ denotes the size of the set $\mathcal{D}$ and $\indicator{\cdot}$ denotes the indicator function.
\end{definition} 

Since $S^{t,s}_{ADE, \epsilon}$ is hard to compute exactly due to the nature of $\bm{x}' \in B_{\epsilon}(\bm{x})$, we approximate it using 50-step PGD attacks, again with a step size of $2.5 \epsilon/50$, maximising the cross entropy loss of the student output's against the hard predictions of its teacher, and instead report \textit{empirical distillation agreement}, $\tilde{\mathcal{A}}^t_{\epsilon}(f_s)$.

\subsection{Results}
\paragraph{Robustness.}
The empirical distillation agreements, \(\tilde{\mathcal{A}}^{RT}_{\epsilon}\), for student networks on MNIST, F-MNIST, and CIFAR-10 datasets are presented in \Cref{tab:robust - robust accuracy - table}. Robust accuracies can be found in the appendix in \Cref{tab: robust accuracies}. We observe, as expected, that adversarial distillation methods (ARD, RSLAD, FD) create more robust students compared to SD, with RSLAD achieving the highest robust accuracy on MNIST and F-MNIST, while FD performs best on CIFAR-10. This highlights that \textit{FD can produce comparably robust students to ARD and RSLAD}. Additionally, adversarial training is shown to better align teacher and student predictions under perturbation, with \(f_{FD}\) generally obtaining the greatest empirical distillation agreement. On F-MNIST, however, \(f_{RSLAD}\) obtains the greatest empirical distillation agreement score, with \(f_{FD}\) surpassing \(f_{ARD}\). The variation across datasets may be attributed to factors such as network sensitivity to hyperparameter choice or the robustness of the teacher network. Future research should explore these underlying differences further.

\begin{table}[t]
\centering
\caption{Distillation Agreement - empirical distillation agreements of students w.r.t. their teachers on MNIST, F-MNIST and CIFAR-10. \textit{Higher is better.}}
\label{tab:robust - robust accuracy - table}
\footnotesize
\renewcommand{\arraystretch}{0.8}
\begin{tabular}{p{0.3cm}w{c}{1cm}w{c}{1.1cm}w{c}{1.1cm}w{c}{1.1cm}w{c}{1.1cm}}
\toprule
&  & \multicolumn{4}{c}{Empirical Distillation Agreement, $\tilde{\mathcal{A}}^{RT}_{ \epsilon}$ (\%)} \\
  &  $\epsilon$ & $f_{SD}$ & $f_{ARD}$ & $f_{RSLAD}$ & $f_{FD}$ \\ 
\midrule
  \multirow{5}{*}{\rotatebox{90}{MNIST}} 
& 0.025   &     $96.3 $     & 96.7     &     ${\bm{97.7}}$    & 97.3                                 \\
& 0.05    &     $95.5 $        & 95.7    &    96.9      & ${\bm{97.0 }}$                              \\
& 0.1          &   $90.2 $     &  94.3  &      95.1      & ${\bm{95.6}}$                                \\
& 0.15    &  82.2              & 92.2      &   92.7     & ${\bm{93.9}}$                                \\
& 0.2     &      68.7              &  90.0      &  90.3    & ${\bm{91.1}}$                               \\ 
\midrule
 \multirow{5}{*}{\rotatebox{90}{F-MNIST}}
& 4/255   &      86.8      & 92.8     &     ${\bm{94.9}}$    & 93.1                                 \\
& 8/255    &      76.4        & 91.7    &    ${\bm{93.1}}$      &  90.3                              \\
& 12/255     &   66.2     &  87.3  &      ${\bm{89.3}}$     &  85.0                                \\
& 16/255    &     54.6              & 80.4      &   ${\bm{85.7}}$     &  81.9                             \\
& 20/255     &      43.6              &  77.6      &  ${\bm{81.7}}$    &  78.8                              \\ 
\midrule
 \multirow{5}{*}{\rotatebox{90}{CIFAR-10}}
& 4/255   &      60.0      & 67.0     &     $76.0$    & $\bm{80.0}$                                 \\
& 8/255    &      42.0        & 61.0    &    $62.0$      &  $\bm{71.0}$                              \\
& 12/255     &   31.0     &  47.0  &      $52.0$     &  $\bm{57.0}$                                \\
& 16/255    &     20.0              & 38.0      &   $44.0$     &  $\bm{55.0}$                             \\
& 20/255     &      13.0              &  31.0      &  ${34.0}$    &  $\bm{43.0}$                              \\ 
\bottomrule
\end{tabular}
\end{table}

\paragraph{Faithfulness and Relative Calibration.}
To understand the faithfulness -- and therefore relative calibration of confidences -- of the different students w.r.t. the robust teacher networks, we report aggregated values of empirical lower bounds (\textsc{EmpLB}) and verified faithfulness bounds (\textsc{FaithUB}) over the test set, and for each $\epsilon$, over all datasets in \Cref{tab: Adv - Faithfulness - Bounds - Avg + STD}. To fully capture the distribution of the bounds over the test set, we additionally present results in \Cref{fig:MNIST histograms} for MNIST and in \Cref{fig:fashion-MNIST histograms} for F-MNIST, both of which can be found in the appendix. For both measures, \textit{lower is better, implying a greater degree of relative calibration between a teacher-student pair.}

\begin{table*}[t]
\centering
\caption{Faithfulness and relative calibration - empirical lower bounds (\textsc{EmpLB}) and faithfulness upper bounds (\textsc{FaithUB}) of students on MNIST, F-MNIST and CIFAR-10. \textit{Lower is better.}}
\label{tab: Adv - Faithfulness - Bounds - Avg + STD}
\renewcommand{\arraystretch}{0.8}
\small
\begin{tabular}{ccw{c}{1.5cm}w{c}{1.5cm}w{c}{1.5cm}w{c}{1.5cm}w{c}{1.5cm}w{c}{1.5cm}w{c}{1.5cm}w{c}{1.5cm}}
\toprule
 & & \multicolumn{4}{c}{\textsc{EmpLB}} & \multicolumn{4}{c}{\textsc{FaithUB}} \\
& $\epsilon$  & $f_{SD}$ & $f_{ARD}$ & $f_{RSLAD}$ & $f_{FD}$  & $f_{SD}$ & $f_{ARD}$ & $f_{RSLAD}$ & $f_{FD}$ \\ 
\midrule
\multirow{5}{*}{\rotatebox{90}{MNIST}} & 0.025              & $0.042_{\pm 0.102}$            &   $0.045_{ \pm 0.113}$  & $0.039_{ \pm 0.087}$ &  $\bm{0.033}_{ \pm 0.079}$   & $0.073_{ \pm 0.150 } $ &     $0.061_{ \pm 0.143}  $   & $0.060_{ \pm 0.118}$         & ${\bm{0.054}_{ \pm 0.113}}$     \\
& 0.05              & $0.060_{ \pm 0.129}$     &   $0.055_{ \pm 0.130 } $ & $0.049_{ \pm 0.101}$ & $\bm{0.041}_{ \pm 0.091}$      &   $0.179_{ \pm 0.257}$  &     $0.097_{ \pm 0.198}$ &  $0.101_{ \pm 0.175 }$ &   ${\bm{0.094}_{ \pm 0.171}}$           \\
& 0.1                & $0.106_{ \pm 0.186}$   &  $0.078_{ \pm 0.164 } $   & $0.072_{ \pm 0.129}$ & $\bm{0.061}_{ \pm 0.117}$ &   $0.731_{ \pm 0.343} $        &  $0.262_{ \pm 0.337}  $  & $0.290_{ \pm 0.319}$               &   ${\bm{0.248}_{ \pm 0.305}}$        \\
& 0.15                & $0.172_{ \pm 0.246}$  &  $0.106_{ \pm 0.197}$ & $0.101_{ \pm 0.160}$  &   $\bm{0.086}_{ \pm 0.145}$      &   $ 0.974_{ \pm 0.123} $         &    $0.636_{ \pm 0.383 } $       &   $0.668_{ \pm 0.355}$        &    ${\bm{0.588}_{ \pm 0.367}}$            \\

& 0.2               & $0.258_{ \pm 0.302}$    &   $0.140_{ \pm 0.229}$ & $0.138_{ \pm 0.192}$ &    $ \bm{0.118}_{ \pm 0.174} $      &   $0.999_{ \pm 0.017} $               &  $0.905_{ \pm 0.230 } $      & $0.889_{ \pm 0.229}  $  &   ${\bm{0.884}_{\pm 0.238}}$           \\ 
\midrule
\multirow{5}{*}{\rotatebox{90}{F-MNIST}} & $4/255$              & $0.099_{\pm 0.159}$            &   $0.067_{ \pm 0.112}$  & $\bm{0.053}_{ \pm 0.094}$ &  $0.060_{ \pm 0.099}$   & $0.215_{ \pm 0.267 } $ &     $0.128_{ \pm 0.182}  $   &  ${\bm{0.118}_{ \pm 0.170}}$         & $0.123_{ \pm 0.171}$     \\
& $8/255$              & $0.155_{ \pm 0.212}$     &   $0.096_{ \pm 0.145 } $ & $\bm{0.081}_{ \pm 0.130}$ & $0.089_{ \pm 0.131}$      &   $0.617_{ \pm 0.393}$  &     $0.330_{ \pm 0.342}$ &  $0.341_{ \pm 0.341 }$ &   ${\bm{0.322}_{ \pm 0.334}}$           \\
& $12/255$                & $0.219_{ \pm 0.265}$   &  $0.129_{ \pm 0.179 } $   & $\bm{0.112}_{ \pm 0.163}$ & $0.122_{ \pm 0.165}$ &   $0.892_{ \pm 0.267} $        &  $0.678_{ \pm 0.381}  $  & $0.690_{ \pm 0.370}$               &   ${\bm{0.668}_{ \pm 0.380}}$        \\
& $16/255$               & $0.297_{ \pm 0.310}$  &  $0.165_{ \pm 0.214}$ & $\bm{0.148}_{ \pm 0.197}$  &   $0.160_{ \pm 0.199}$      &   $ 0.990_{ \pm 0.071} $         &    $0.906_{ \pm 0.235 } $       &   $0.917_{ \pm 0.214}$        &    ${\bm{0.902}_{ \pm 0.237}}$            \\

& $20/255$               & $0.378_{ \pm 0.343}$    &   $0.204_{ \pm 0.243}$ & $\bm{0.190}_{ \pm 0.226}$ &    $ 0.202_{ \pm 0.229} $      &   $ 0.999_{\pm 0.013}$               &  $0.986_{ \pm 0.089 } $      & $0.988_{ \pm 0.079}  $  &   ${\bm{0.985}_{\pm 0.090}}$           \\ 
\midrule
\multirow{5}{*}{\rotatebox{90}{CIFAR-10}} & $4/255$              & $0.282_{\pm 0.145}$            &   $0.242_{ \pm 0.139}$  & $0.226_{ \pm 0.126}$ &  $\bm{0.187}_{ \pm 0.096}$   & $0.671_{ \pm 0.191 } $ &     $0.546_{ \pm 0.203}  $   &  $0.497_{ \pm 0.176}$         & ${\bm{0.474}_{ \pm 0.182}}$     \\
& $8/255$              & $0.388_{ \pm 0.166}$     &   $0.317_{ \pm 0.155 } $ & $0.291_{ \pm 0.139}$ & $\bm{0.245}_{ \pm 0.114}$      &   $0.983_{ \pm 0.033}$  &     $0.910_{ \pm 0.127}$ &  $0.873_{ \pm 0.133 }$ &   ${\bm{0.864}_{ \pm 0.153}}$           \\
& $12/255$                & $0.489_{ \pm 0.175}$   &  $0.394_{ \pm 0.169 } $   & $0.357_{ \pm 0.148}$ & $\bm{0.302}_{ \pm 0.129}$ &   $1.000_{ \pm 0.002 } $        &  $1.000_{ \pm 0.016 }  $  & $0.990_{ \pm 0.035 }$               &   ${\bm{0.989}_{ \pm 0.038}} $        \\
& $16/255$               & $0.583_{ \pm 0.177}$  &  $0.462_{ \pm 0.178}$ & $0.415_{ \pm 0.157}$  &   $\bm{0.359}_{ \pm 0.142}$      &   $ 1.000_{\pm 0.000} $         &    $1.000_{ \pm  0.002} $       &   ${\bm{0.999}}_{ \pm 0.008}$        &    ${\bm{0.999}_{ \pm 0.009}}$            \\
& $20/255$               & $0.666_{ \pm 0.165}$  &  $0.530_{ \pm 0.176}$ & $0.474_{ \pm 0.164}$  &   $\bm{0.416}_{ \pm 0.151}$      &   $ 1.000_{\pm 0.000} $         &    $1.000_{ \pm  0.000} $       &   ${\bm{1.000}}_{ \pm 0.001}$        &    ${\bm{1.000}_{ \pm 0.001}}$            \\
\bottomrule
\end{tabular}
\end{table*}

From \Cref{tab: Adv - Faithfulness - Bounds - Avg + STD}, we observe that the empirical attack bounds (\textsc{EmpLB}) are on average smaller for $f_{FD}$, $f_{RSLAD}$ and $f_{ARD}$ than for $f_{SD}$. This indicates that the adversarially distilled students are, on average, \textit{empirically more relatively well-calibrated than SD students} w.r.t. the robust teacher. This is exemplified further in \Cref{fig:MNIST histograms}, where there is a greater shift in the empirical bound distributions for $f_{SD}$ over $f_{FD}$, $f_{RSLAD}$ and $f_{ARD}$. In addition, on the MNIST and CIFAR-10 datasets, we see that the empirical bounds are lower for $f_{FD}$ than for the three other student networks. Moreover, on MNIST and CIFAR-10, we notice the standard deviations of the bounds for $f_{FD}$ are smaller across all values of $\epsilon$, showing a more concentrated spread of lower empirical bounds than for the other student, with fewer images creating large differences in the confidences of $f_{FD}$ and $f_{RT}$. This is further highlighted in the upper tails of the \textsc{EmpLB} distributions in \Cref{fig:MNIST histograms}. On F-MNIST, like in \Cref{tab:robust - robust accuracy - table}, we again observe a change of results, with $f_{RSLAD}$ attaining the lowest empirical bounds followed by $f_{FD}$.

Looking at the faithfulness bounds (\textsc{FaithUB}) for the $f_{FD}$, $f_{RSLAD}$ and $f_{ARD}$ students on both datasets in \Cref{tab: Adv - Faithfulness - Bounds - Avg + STD}, we observe that they are significantly lower across all values of $\epsilon$ than for $f_{SD}$, which confirms the empirical observations discussed above. Moreover, the higher standard deviation of the faithfulness bounds for $\epsilon$ values of $0.15$ and $0.2$ for $f_{FD}$, $f_{SD}$ and $f_{ARD}$ indicate that these students have less of an accumulation of images with faithfulness bounds of $1$ - the largest possible difference in confidences between a teacher and its student. This is further shown on MNIST in 
\Cref{fig:MNIST histograms}, where we observe empirically and verifiably that the maximum confidence difference between the adversarially distilled students and their teacher is smaller for a greater number of images, with their faithfulness bounds more closely imitating the empirical attack bounds for smaller values of $\epsilon$ than for $f_{SD}$. This suggests, in particular, that the \textit{degree of relative calibration between these adversarially trained students and their teacher is greater in a verifiable, upper-bound sense}. 

\begin{figure}
    \centering
    \includegraphics[width=0.75\linewidth]{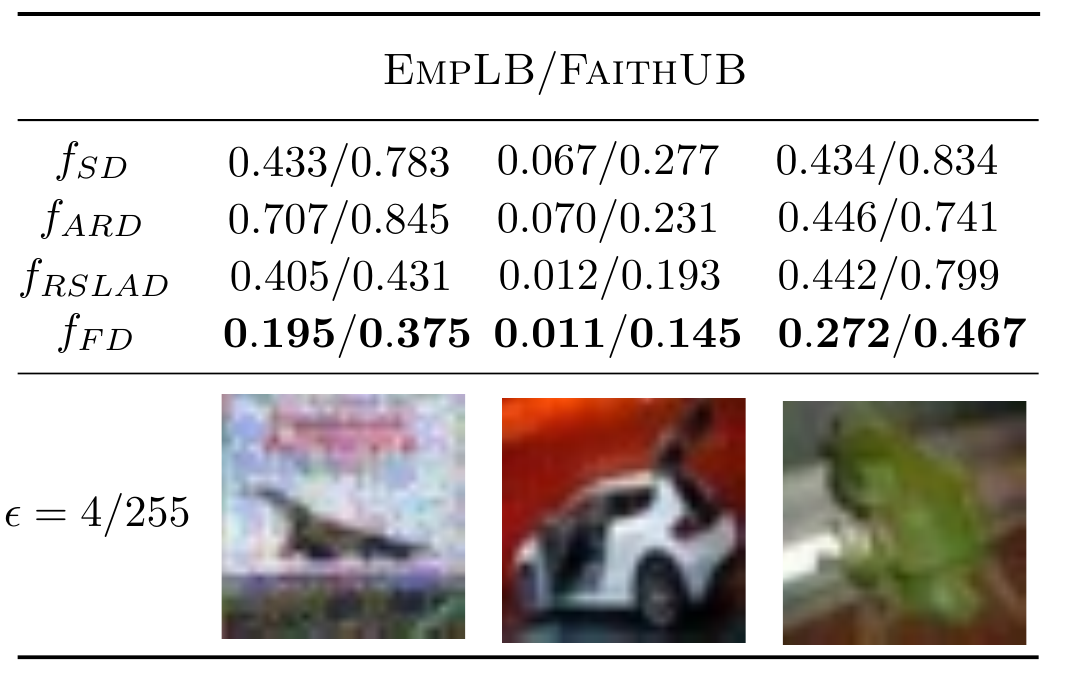}
    \caption{CIFAR-10 \textsc{EmpLB} and \textsc{FaithUB} - examples of perturbed images with $\epsilon = 4/255$ from the CIFAR-10 test set alongside the \textsc{EmpLB}s and \textsc{FaithUB}s of student networks for these images.}
    \label{fig:enter-label}
\end{figure}

On all datasets, we see that for nearly all values of $\epsilon$, we observe that the faithfulness bounds are, on average lower for $f_{FD}$ than for all of the other students across both datasets. \textit{This shows that $f_{FD}$ is verifiably more relatively calibrated in an upper bound sense to its teacher on both MNIST, F-MNIST and CIFAR-10.} On MNIST and CIFAR-10, this aligns and supports the empirical observations that $f_{FD}$ is more faithful to its teacher in terms of confidences. 

On F-MNIST, we have a disparity where $f_{FD}$ is verifiably more relatively calibrated across nearly all of the values of $\epsilon$ as shown by lower verified bounds but empirically is observed to be less well relatively calibrated to its teacher than $f_{RSLAD}$, as shown by $f_{FD}$'s greater empirical faithfulness bounds and lower empirical distillation agreement.

Finally, we note that the standard deviations of the computed $\textsc{FaithUB}$ on all datasets tend to be large. This results from the fact that the bounds are bounded above by 1. In particular, this means that high variance is desirable in cases where the mean is higher. This is observable from the distribution of the MNIST bounds for $\epsilon = 0.15$ and $0.2$ plotted in \Cref{fig:MNIST histograms}, where we notice that the higher standard deviation of FD comes from having a greater accumulation of bounds in lower sections of the histograms.

\paragraph{FD as a more verifiable distillation method over RSLAD.} 
\label{sec: FD vs RSLAD}
As mentioned above, from \Cref{tab:robust - robust accuracy - table} and \Cref{tab: Adv - Faithfulness - Bounds - Avg + STD}, we see that the empirical measures of distillation robustness that we have introduced, that of $\textsc{EmpLB}$ and empirical distillation agreement, are worse for $f_{FD}$ than for $f_{RSLAD}$ on F-MNIST. However, we note that the general trend of $f_{FD}$ producing tighter $\textsc{FaithUB}$ holds for all datasets.

These empirical measures, however, provide no guarantees on the relative calibration of a student to its teacher. Indeed, such methods are highly dependent on the method of attack used for their computation. In particular, \Cref{tab: FD vs RLSAD}, which can be found in the appendix, shows three examples, $\bm{x}_1$, $\bm{x}_2$, and $\bm{x}_3$, from the F-MNIST test set where, for $\epsilon = 8/255$, we compute $\textsc{EmpLB}$s using PGD attacks with steps of $1, 10, 25$, and $50$. We observe that $f_{RSLAD}$ and $f_{FD}$ flip between giving the lowest and hence the best $\textsc{EmpLB}$. This indicates that the aforementioned empirical methods cannot solely be used for comparisons. Combined with the difference in results on MNIST and CIFAR-10 to F-MNIST, this highlights the importance of verifiable methods of bounding the maximum difference in confidences between a teacher and student. In particular, $\textsc{FatihUB}$s provide certified upper bounds as they are computed using linear relaxations as a MILP and are therefore independent of and not subject to a choice of attack method such as PGD. As a result, $\textsc{FaithUB}$s are a more robust method of evaluating the relative calibration of a teacher-student pair and, importantly, provide guarantees for downstream applications such as in safety-critical devices. It is still worth noting that a more complete picture is given when $\textsc{FaithUB}$ is evaluated alongside $\textsc{EmpLB}$.

\subsection{Limitations} For larger values of $\epsilon$, the verified faithfulness bounds for all our student networks are looser compared to the empirically produced bounds. This indicates that our method of computing faithfulness bounds struggles to scale with increasing values of $\epsilon$. Since LP methods will provide tighter bounds than bound propagation methods such as CROWN \citep{zhang2018efficient}, this proves to be a limitation for verifying the greater degree of relative calibration seen empirically for larger values of $\epsilon$. This alludes to the fact that a more sophisticated method of producing faithfulness bounds needs to be produced to verify larger networks for larger values of $\epsilon$ on more complicated data sets. Nevertheless, the framework and methods we introduce in this work serve as a solid foundation for future investigation into the confident deployment of robust student networks, a consideration paramount in safety-critical environments.

\section{Conclusion}
The setting of faithful imitators provides a framework for empirically and verifiably reasoning about the relative calibration of confidences, that is, the maximum difference in confidences, of a teacher-student pair in a KD setting. This analysis is essential for the safe deployment of student networks in safety-critical environments. Our experiments on MNIST, Fashion-MNIST and CIFAR-10 suggest that when combined with a robust teacher, ARD, RSLAD, and our FD training produce empirically and verifiably better relatively-calibrated teacher-student network pairs than non-adversarial distillation, with FD proving to be a viable alternative method of adversarial distillation. Moreover, we find that FD-trained students are verifiably more relatively well-calibrated to their teacher network. However, we observe that the results of our empirical methods vary across datasets, highlighting the need for verifiable guarantees provided by our framework and methods. Future work should further explore the relative calibration between teacher-student pairs, including the disparity observed between empirical and verified methods on larger datasets, applying and further adapting the novel framework that we have introduced in this work. 

\bibliography{aaai24}

\appendix
\section{Experiment Details}

\begin{table*}[t]
\renewcommand{\arraystretch}{0.8}
\centering
\small
\caption{CIFAR-10 architectures - CIFAR-10 teacher and student network architectures.}
\label{tab:CIFAR-10 network_architectures}
\begin{tabular}{cc}
\toprule
\textbf{Teacher Network} & \textbf{Student Networks} \\
\midrule
Conv(input=3, filters=8, kernel=3, stride=2) & Conv(input=3, filters=6, kernel=3, stride=2) \\
ReLU & ReLU \\
Pool(2, stride=2) & Pool(2, stride=2) \\
Conv(input=8, filters=16, kernel=3, stride=2, pad.=1) & Conv(input=6, filters=16, kernel=3, stride=2, pad.=1) \\
ReLU & ReLU \\
Conv(input=16, filters=32, kernel=3, stride=1) & Conv(input=16, filters=32, kernel=3, stride=1) \\
FC(input=128, output=128) & FC(input=128, output=64) \\
ReLU & ReLU \\
FC(input=128, output=64) & FC(input=64, output=32) \\
ReLU & ReLU \\
FC(input=64, output=10) & FC(input=32, output=10) \\
\bottomrule
\end{tabular}
\end{table*}

\subsection{Training Settings for Networks Used in MNIST Experiments}
\label{MNIST training}
We trained all models over 64 epochs with early stopping (patience of 8 epochs) based on test accuracy. Teacher and student networks have \{30, 30, 30, 30, 30\} and \{20, 20, 20, 20\} neurons per layer, respectively. Students were trained using SD ($f_{SD}$), ARD ($f_{ARD}$), RSLAD ($f_{RSLAD}$), and our loss, FD ($f_{FD}$), achieving test accuracies of $97.41\%$, $97.20\%$, $97.12\%$, and $97.14\%$ on MNIST, respectively.

The teacher, $f_{RT}$, was trained with adversarial training, using PGD attacks (\(\epsilon = 0.2\), step size \(0.05\), 10 iterations), SGD with momentum of $0.9$, weight decay $0.002$, and initial learning rate of $0.04$, decaying to $0.04/32$ with cosine annealing.

Student training configurations were similar across methods. For $f_{SD}$, we used SGD (momentum $0.9$, temperature $4$, weight decay $0.001$, learning rate from $0.04$ to $0.04/16$), and \(\alpha = 0.5\). For $f_{ARD}$, $f_{RSLAD}$, and $f_{FD}$, training included PGD (10 steps, step size $0.05$, \(\epsilon = 0.2\)), temperature $2$, weight decay $0.001$, and \(\alpha = 0.5\). Learning rates were $0.04$ decaying to $0.04/32$ for $f_{ARD}$ and $f_{RSLAD}$, and from $0.04$ to $0.04/16$ for $f_{FD}$. Again, we use cosine annealing for all learning rate decays.

\subsection{Training Settings for Networks Used in F-MNIST Experiments}
\label{Fashion - MNIST training}
We train all models over 64 epochs with early stopping (patience of 8 epochs), using teacher and student networks with neurons \{64, 32, 32, 32, 32, 16, 10\} and \{30, 30, 30, 30, 10\} respectively. The clean test set accuracies on F-MNIST are \(88.29\%\), \(88.18\%\), \(87.47\%\), and \(87.81\%\) for students using SD ($f_{SD}$), ARD ($f_{ARD}$), RSLAD ($f_{RSLAD}$), and our loss, FD ($f_{FD}$). The teacher, $f_{RT}$, was trained using PGD attacks (\(\epsilon = 8/255\), step size \(2/255\), 10 iterations), SGD (momentum 0.9, weight decay 0.002, learning rate from 0.04 to 0.04/32). 

Student $f_{SD}$ was trained with a temperature of 4, weight decay 0, learning rate from 0.01 to 0.01/32; $f_{ARD}$ with weight decay 0.001, learning rate from 0.01 to 0.01/8; $f_{RSLAD}$ with weight decay 0, learning rate from 0.02 to 0.02/32; and $f_{FD}$ with weight decay 0.001, learning rate from 0.02 to 0.02/32. All students were trained using the PGD (\(\epsilon = 8/255\), step size \(2/255\), 10 steps), SGD (momentum 0.9), temperature 2, loss mixing weight \(\alpha = 0.5\). Again, we use cosine annealing for all learning rate decays.

\subsection{Training Settings for Networks Used in CIFAR-10 Experiments}
\label{CIFAR training}
We train all models over 128 epochs with early stopping (patience of 12 epochs) based on test accuracy; see \Cref{tab:CIFAR-10 network_architectures} for architectures. Students using SD ($f_{SD}$), ARD ($f_{ARD}$), RSLAD ($f_{RSLAD}$), and FD ($f_{FD}$) attained clean test set accuracies of \(63.09\%\), \(62.28\%\), \(60.55\%\), \(60.48\%\) on CIFAR-10.

The teacher $f_{RT}$ was trained using PGD attacks (\(\epsilon = 8/255\), step size \(2/255\), 10 iterations), SGD (momentum 0.9, weight decay 0.001, learning rate from 0.1 to 0.01/32).

All students were trained with the same PGD and SGD parameters, temperature 2, one run of cosine annealing, and using the following settings: $f_{SD}$ with weight decay 0.001, learning rate from 0.1 to 0.1/32, \(\alpha = 0.5\); $f_{ARD}$ with weight decay 0.001, learning rate from 0.1 to 0.1/32, \(\alpha = 0.5\); $f_{RSLAD}$ with weight decay 0, learning rate from 0.1 to 0.1/16, \(\alpha = 0.5\); $f_{FD}$ with weight decay 0, learning rate from 0.1 to 0.1/32, \(\alpha = 1.0\) (\Cref{tab: ablation study} shows an ablation study of bounding results while varying \(\alpha\)). Again, we use cosine annealing for all learning rate decays.

\subsection{Motivation for Verified Measures of Faithfulness}

\begin{table}[h]
\centering
\caption{Lack of guarantees of $\textsc{EmpLB}$ -  $\textsc{EmpLB}$ values computed using PGD attacks with varying numbers of optimisation steps for three examples, $\bm{x}_1$, $\bm{x}_2$ and $\bm{x}_3$, from the F-MNIST test set. }
\label{tab: FD vs RLSAD}
\footnotesize
\begin{tabular}{ccw{c}{1.0cm}w{c}{1.0cm}w{c}{1.0cm}w{c}{1.0cm}} \toprule 
                      &       & \multicolumn{4}{c}{EmpLB}             \\
                      &       & PGD-1   & PGD-5   & PGD-25  & PGD-50  \\
\midrule
\multirow{2}{*}{$\bm{x}_1$} & RSLAD & $\bm{0.04679}$ & $\bm{0.06828}$ & 0.07177 & 0.07182 \\
                      & FD    & 0.05060 & 0.06968 & $\bm{0.07091}$ & $\bm{0.07111}$ \\
\midrule
\multirow{2}{*}{$\bm{x}_2$} & RSLAD & $\bm{0.1221}$  & 0.1401  & $\bm{0.1420}$  & $\bm{0.1426}$  \\
                      & FD    & 0.1240  & $\bm{0.1381}$  & 0.1431  & 0.1439  \\
\midrule
\multirow{2}{*}{$\bm{x}_3$} & RSLAD & $\bm{0.1289}$  & $\bm{0.1821}$  & $\bm{0.1861}$  & 0.1884  \\
                      & FD    & 0.1879  & 0.1879  & 0.1879  & $\bm{0.1879}$ \\
\bottomrule
\end{tabular}
\end{table}

\begin{table*}[h]
\renewcommand{\arraystretch}{0.8}
\caption{CIFAR-10 mixing weight ablation - ablation study investigating the effect of changing the mixing weight, that is $\alpha$ in \Cref{FD loss}, during training on the values $\textsc{EmpLB}$ and $\textsc{FaithUB}$.}
\label{tab: ablation study}
\begin{tabular*}{\textwidth}{@{\extracolsep{\fill}} cccccccc}
\toprule
 & & \multicolumn{3}{c}{\textsc{EmpLB}} & \multicolumn{3}{c}{\textsc{FaithUB}} \\
& $\epsilon$  & $f_{FD, 0.5}$ & $f_{FD, 0.75}$ & $f_{FD, 1.0}$ & $f_{FD, 0.5}$  & $f_{FD, 0.75}$ & $f_{FD, 1.0}$   \\ 
\midrule
\multirow{5}{*}{\rotatebox{90}{CIFAR-10}} & $4/255$              & $0.199_{\pm 0.119}$            &   $0.212_{ \pm 0.120}$  &  $\bm{0.187}_{ \pm 0.096}$   & $0.525_{ \pm 0.193 } $ &     $0.511_{ \pm 0.196}  $    & ${\bm{0.474}_{ \pm 0.182}}$\\
& $8/255$              & $0.265_{ \pm 0.134}$     &   $0.273_{ \pm 0.137 } $ &  $\bm{0.245}_{ \pm 0.114}$      &   $0.907_{ \pm 0.118}$  &     $0.889_{ \pm 0.132}$      &   ${\bm{0.864}_{ \pm 0.153}}$  \\

& $12/255$                & $0.333_{ \pm 0.145}$   &  $0.333_{ \pm 0.150 } $   &  $\bm{0.302}_{ \pm 0.129}$ &   $0.993_{ \pm 0.030 } $        &  $0.991_{\pm 0.035 }  $     &  ${\bm{0.989}_{ \pm 0.038}}$ \\
& $16/255$               & $0.400_{ \pm 0.150}$  &  $0.394_{ \pm 0.159}$ &   $\bm{0.359}_{ \pm 0.142}$      &   $ 0.999_{ \pm 0.008} $         &    $0.999_{ \pm  0.008} $            &  ${\bm{0.999}_{ \pm 0.009}}$  \\
& $20/255$               & $0.462_{ \pm 0.153}$  &  $0.451_{ \pm 0.166}$ &   $\bm{0.416}_{ \pm 0.151}$      &   $ {1.000 \pm 0.000} $         &    $1.000_{ \pm  0.001} $            &  ${\bm{1.000}_{ \pm 0.001}}$  \\
\bottomrule
\end{tabular*}
\end{table*}
\section{Experiment Results}
\subsection{Ablation of Loss Weighting $\alpha$ for FD on CIFAR-10}
\label{Ablation}
Table \ref{tab: ablation study} shows an ablation study carried out to investigate the effect of the loss weighting $\alpha$ on the values $\textsc{EmpLB}$ and $\textsc{FaithUB}$. Here, $f_{FD, \alpha}$ denoted a distilled FD with mixture weighting $\alpha$. Here, we see a clear trend in that increasing the weight of the newly introduced FD term during training leads to increased tighter empirical and verified faithfulness bounds.

\subsection{Robust Accuracies of Models}
\label{sec: robust accurcies}
\begin{table}[ht]
\centering
\caption{Robustness - robust accuracies of trained networks on MNIST, F-MNIST and CIFAR-10.}
\label{tab: robust accuracies}
\renewcommand{\arraystretch}{0.8}
\footnotesize
\begin{tabular}{p{0.01cm}cccccc}
\toprule
& & \multicolumn{5}{c}{Individual Model Robust Accuracy (\%)}  \\
  &  $\epsilon$ & $f_{RT}$ & $f_{SD}$ & $f_{ARD}$ & $f_{RSLAD}$ & $f_{FD}$   \\ 
\midrule
  \multirow{6}{*}{\rotatebox{90}{MNIST}} & 0       & 97.9                    & 97.3            &    96.6      & 97.0                       & 97.0                              \\
& 0.025   & 97.5                    & 95.1            &      95.9     & $\bm{96.5}$                       & 95.9                                  \\
& 0.05    & 96.7                    & 94.1            &     94.7      & $\bm{95.7}$                       & 95.5             \\
& 0.1     & 94.9                    & 88.7           &      93.0     &  $\bm{93.8}$                       &   93.4       \\
& 0.15    & 92.9                    & 80.3            &      89.9     &  $\bm{90.8 }$                      &  $\bm{90.8 }$   \\
& 0.2     & 90.0                    & 65.6           &      86.7     & ${\bm{87.0}}$                       &  86.0          \\ 
\midrule
 \multirow{6}{*}{\rotatebox{90}{F-MNIST}} & 0       & 89.1                    & 89.1            &    89.1      & 88.4          & 88.6         \\
& 4/255   & 86.5                    & 81.2            &      85.5     & ${\bm{85.9}}$                       & 85.5         \\
& 8/255    & 83.8                    & 72.3            &     82.9      &  ${\bm{83.5}}$                      &  82.4         \\
& 12/255     & 80.9                    & 61.2            &      78.2     &  ${\bm{79.0}}$                      &   77.0      \\
& 16/255    & 76.9                    &  49.1            &      72.2     &  $\bm{74.5}$                      &  72.2          \\
& 20/255     & 72.2                    & 37.7           &      67.3     & ${\bm{68.6}}$                       &  67.1                 \\ 
\midrule
 \multirow{6}{*}{\rotatebox{90}{CIFAR-10}} & 0       & 72.0                    & 70.0            &    68.0      & 65.0          & 74.0         \\
& 4/255   & 68.0                    & 53.0            &      60.0     & 59.0                      & $\bm{69.0}$         \\
& 8/255    & 58.0                    & 36.0            &     53.0      &  $48.0$                      &  $\bm{56.0}$         \\
& 12/255     & 46.0                    & 27.0            &      40.0     &  ${38.0}$                      &   $\bm{42.0}$      \\
& 16/255    & 35.0                    &  14.0            &      29.0     &  $30.0$                      &  $\bm{37.0}$          \\
& 20/255     & 27.0                    & 5.0           &      21.0     & $21.0$                       &  $\bm{25.0}$                 \\ 
\bottomrule
\end{tabular}
\end{table}

\begin{figure*}[h]
\centering
\includegraphics[width=0.7\linewidth]{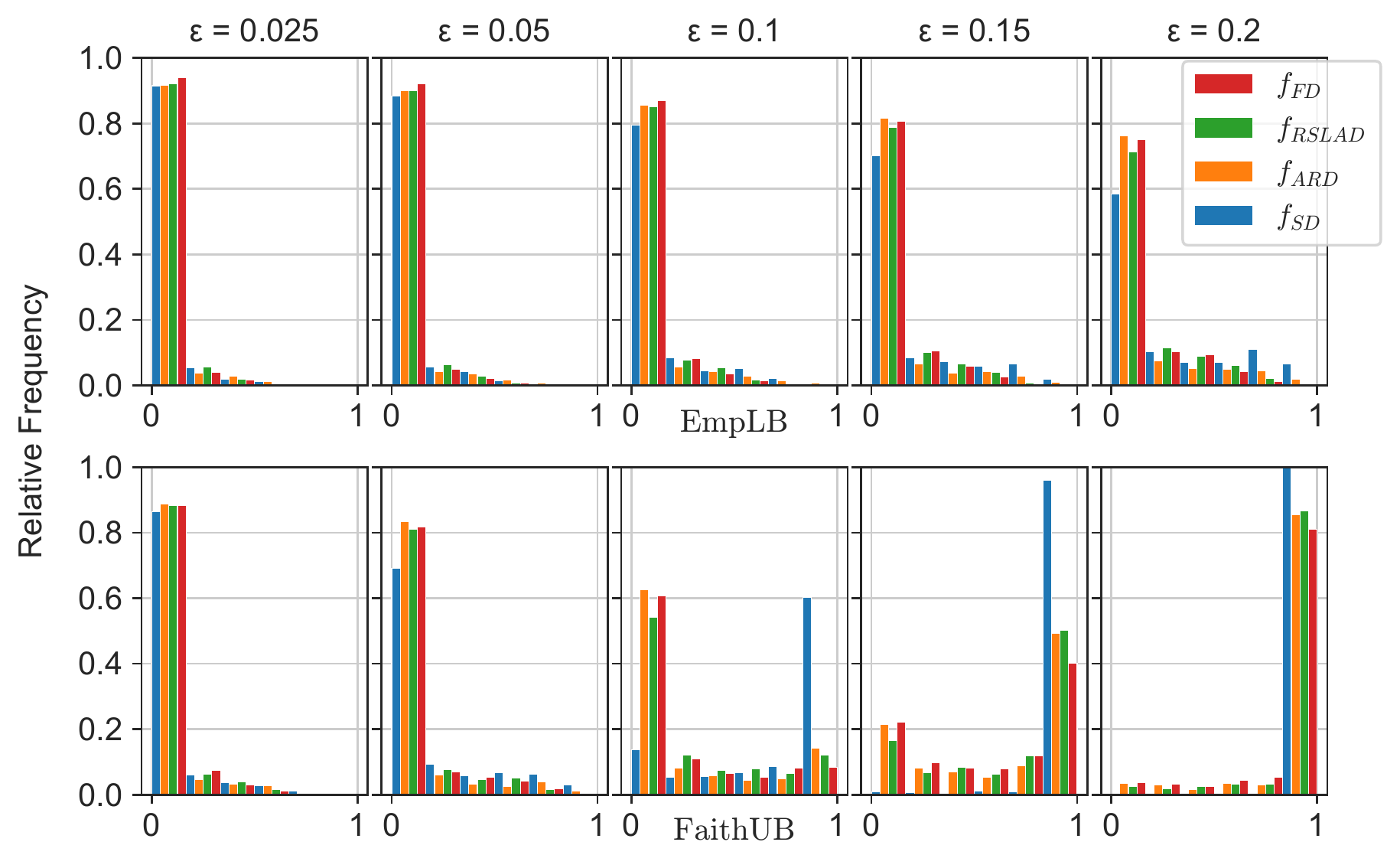}
\caption{MNIST Faithfulness and Relative Calibration - histograms of the distributions of empirical lower bounds (\textsc{EmpLB}) and faithfulness upper bounds (\textsc{FaithUB}) over the test set for different values of $\epsilon$, for students distilled from the robust teacher ($f_{RT}$) using standard distillation ($f_{SD}$) - left bar, ARD ($f_{SD}$) - left-middle bar, RSLAD ($f_{RSLAD}$) - right-middle bar and FD ($f_{FD}$) - right bar on the MNIST data set.}
\label{fig:MNIST histograms}
\end{figure*}

\begin{figure*}[h]
\centering
\includegraphics[width=0.7\linewidth]{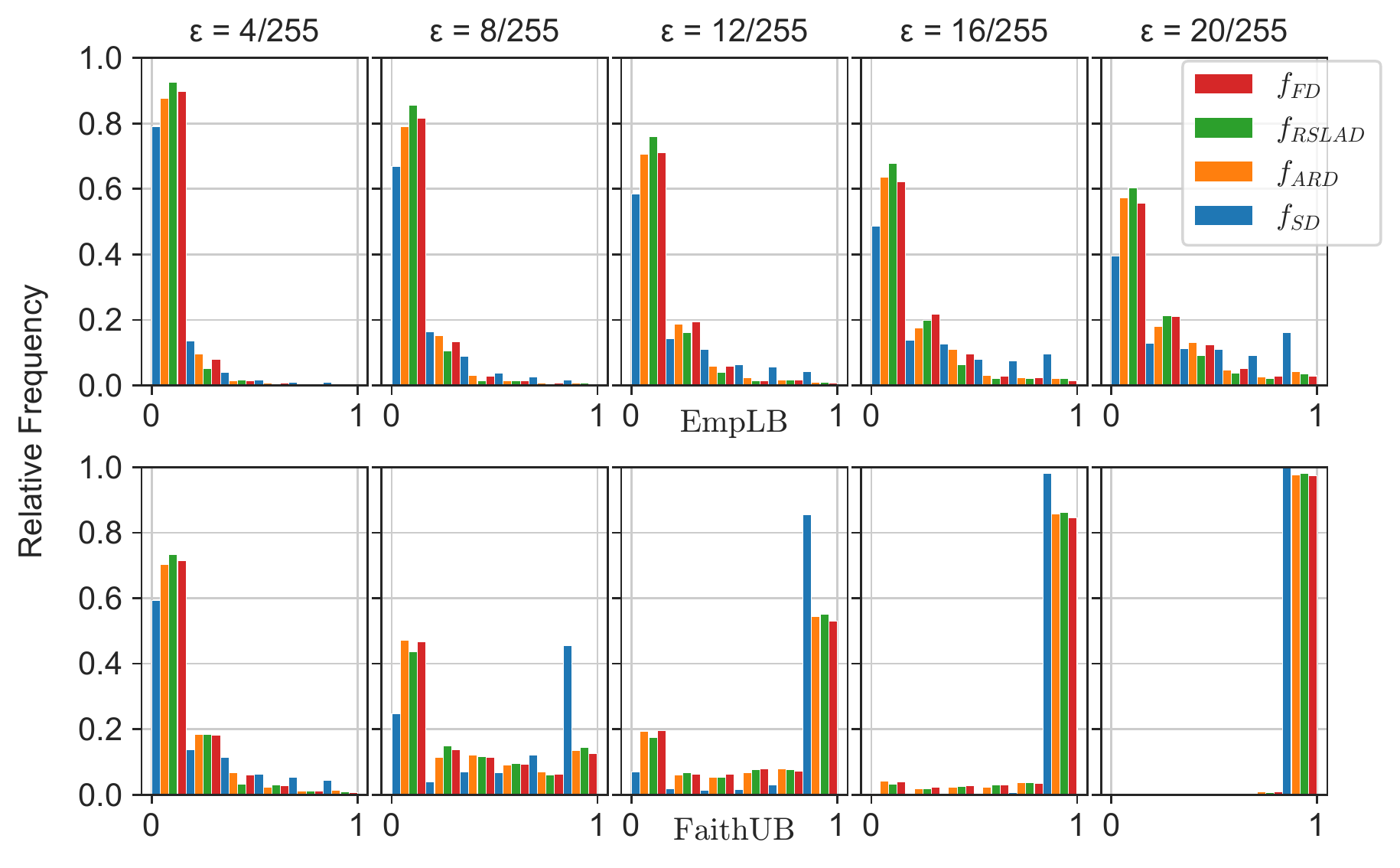}
\caption{F-MNIST Faithfulness and Relative Calibration - histograms of the distributions of empirical lower bounds (\textsc{EmpLB}) and faithfulness upper bounds (\textsc{FaithUB}) over the test set for different values of $\epsilon$, for students distilled from the robust teacher ($f_{RT}$) using standard distillation ($f_{SD}$) - left bar, ARD ($f_{SD}$) - left-middle bar, RSLAD ($f_{RSLAD}$) - right-middle bar and FD ($f_{FD}$) - right bar on the F-MNIST data set.}
\label{fig:fashion-MNIST histograms}
\end{figure*}

\end{document}